\documentclass{article}
\usepackage{spconf,amsmath,epsfig}
\usepackage{algorithm,algorithmic}
\usepackage{graphicx}

\title{ADDRESSING SINGLE OBJECT TRACKING IN SATELLITE IMAGERY THROUGH PROMPT-ENGINEERED SOLUTIONS}
\fourauthors
  {Athena Psalta}
  {RSLab\\	NTUA\\	Athens, Greece}
  {Vasileios Tsironis}
  {RSLab\\	NTUA\\	Athens, Greece}
  {Andreas El-Saer}
  {Photogrammetry Lab\\	UniWA\\	Athens, Greece}
  {Konstantinos Karantzalos}
  {RSLab\\	NTUA\\	Athens, Greece}

\begin{document}
%
\maketitle
\begin{abstract}
Object tracking in satellite videos remains a complex endeavor in remote sensing due to the intricate and dynamic nature of satellite imagery. Existing state-of-the-art trackers in computer vision integrate sophisticated architectures, attention mechanisms, and multi-modal fusion to enhance tracking accuracy across diverse environments. However, the challenges posed by satellite imagery, such as background variations, atmospheric disturbances, and low-resolution object delineation, significantly impede the precision and reliability of traditional Single Object Tracking (SOT) techniques. Our study delves into these challenges and proposes prompt engineering methodologies, leveraging the Segment Anything Model (SAM) and TAPIR (Tracking Any Point with per-frame Initialization and temporal Refinement), to create a training-free point-based tracking method for small-scale objects on satellite videos. Experiments on the VISO dataset validate our strategy, marking a significant advancement in robust tracking solutions tailored for satellite imagery in remote sensing applications.

\end{abstract}
\section{Introduction}
\label{introduction}

Single Object Tracking (SOT) in satellite videos stands as a critical yet intricate task in remote sensing due to the complex and dynamic nature of satellite imagery. Precisely tracing and defining specific objects within these videos holds immense significance across diverse applications, ranging from environmental observation to urban development and security monitoring \cite{urban_development}. Current state-of-the-art trackers \cite{tracking_every} \cite{aiatrack} in computer vision integrate complex architectures \cite{Cheng_2023_ICCV}, attention mechanisms \cite{Cai_2023_ICCV} and multi-modal fusion \cite{track_multimodal}, enhancing accuracy and adaptability across diverse environments. 

Algorithms for SOT typically fall into two main categories: bounding box-based and point-based trackers. Bounding box-based trackers enclose the target object within a rectangular bounding box and focus on predicting and adjusting the box's position, scale, and orientation across frames. However, in the context of satellite videos capturing small and densely moving objects, the traditional use of bounding box-based trackers is less effective due to the minute size of the targets. Objects in satellite imagery often require more nuanced and precise tracking methodologies, rendering point-based trackers more suitable. These point-based trackers excel in capturing and tracing small-scale objects by focusing on individual points or features, offering enhanced adaptability and accuracy in discerning and consistently tracking amidst diverse and unpredictable backgrounds.

However, the distinctive challenges inherent in satellite imagery, notably the pervasive presence of background variations, atmospheric disturbances, varying lighting conditions, and scale discrepancies, significantly impede the precision and reliability of traditional SOT techniques. Conventional point-based trackers, while robust in controlled environments, encounter substantial difficulties in discerning and consistently tracking objects amidst the diverse and often unpredictable backgrounds captured in satellite footage. These inherent complexities underscore the pressing need for innovative approaches that transcend the limitations of conventional tracking algorithms and harness novel methodologies, such as prompt engineering, to achieve more resilient and accurate object tracking in satellite videos. 

\begin{figure*}
  \includegraphics[width=\textwidth, height=8.5cm]{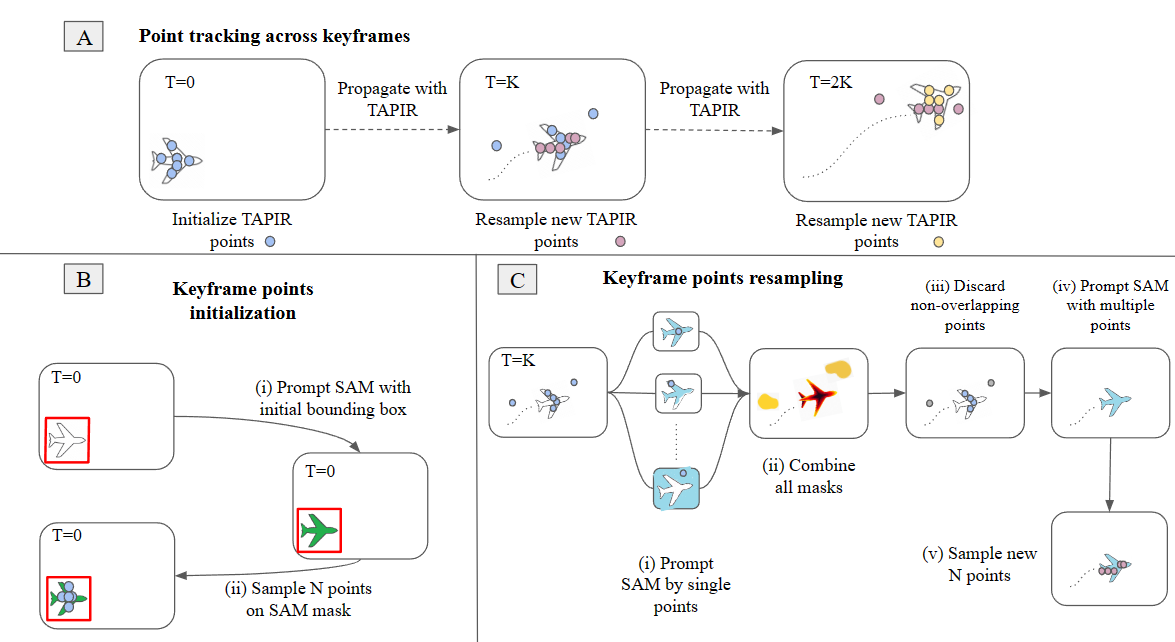}
  \caption{\textbf{Methodology overview of our SOT pipeline}. A) demonstrates the continuous tracking of points of interest, while B) and C) highlight the essential steps of initializing and sampling points during keyframe transitions, forming a comprehensive strategy for effective SOT in satellite videos.}
  \label{method-overview}
\end{figure*}

Existing methods for object tracking in satellite imagery often grapple with these challenges by employing adaptations of generic visual tracking algorithms. Some approaches focus on refining traditional methods, such as using motion models or region-based tracking, to accommodate the complexities of satellite imagery. Others have attempted to integrate machine learning techniques, although these methods commonly face limitations when dealing with the low-resolution, sparse, and diverse nature of objects within satellite videos. The lack of robustness in these approaches highlights the need for innovative strategies capable of handling the intricate and variable backgrounds while effectively tracking objects.

To this end, our approach is motivated by the inadequacies of current methods in addressing the object tracking complexities within satellite videos, aiming to leverage prompt engineering techniques to overcome these limitations and advance the precision and adaptability of tracking methodologies in this domain. Integrating the strengths of the Segment Anything Model (SAM) \cite{sam} and TAPIR (Tracking Any Point with per-frame Initialization and temporal Refinement) \cite{tapir}, our approach aims to navigate background variations and low-resolution object delineation, enhancing the adaptability and efficacy of object tracking. SAM, known for its robust segmentation capabilities, accurately delineates objects of interest, providing precise initial inputs for TAPIR. Leveraging TAPIR's strength in per-frame initialization and temporal refinement, our approach navigates complexities like background variations and low-resolution object details, enhancing the adaptability and efficacy of tracking. By synergizing these techniques, our approach optimizes object delineation and tracking, enhancing adaptability and robustness within the intricate landscape of satellite imagery. 

Our contribution lies in offering a refined methodology within satellite imagery based on a prompt-engineering approach, bypassing the need for training by utilizing pre-trained SAM and TAPIR. This strategy enhances adaptability, allowing seamless integration into diverse tracking scenarios in satellite videos. We conducted experiments on the VISO dataset \cite{viso}, showcasing promising outcomes. These results validate the efficacy of the prompt-engineering methodology, demonstrating robust tracking performance in challenging satellite imagery environments.

\section{METHODOLOGY}
\label{methodology}

Our proposed SOT methodology leverages the TAPIR point tracker and the SAM segmentation model, as shown in Figure \ref{method-overview}. The tracking pipeline operates on keyframes, each separated by a fixed interval K. TAPIR tracks a set of points between keyframes, while a SAM-based strategy renews them at each keyframe. The initial keyframe employs SAM for precise point initialization based on an initial bounding box for the target. This structured approach ensures accurate and adaptive tracking, optimizing the interplay between TAPIR and SAM throughout the sequence. In the subsequent sections, we delve deeper into the intricacies of our methodology, while also providing a comprehensive analysis of the underlying algorithms, SAM and TAPIR.

\subsection{Pre-trained models}

The Segment Anything Model (SAM) plays a pivotal role in our methodology for SOT in satellite videos. SAM \cite{sam} operates as a segmentation-based model designed to identify and segment objects of interest within video frames. SAM accepts various prompt inputs, including bounding boxes, individual points, or textual descriptions. For our approach, we specifically utilize bounding boxes as the input prompt for SAM.
This algorithm starts by receiving an initial bounding box or region of interest (ROI) containing the object in the first frame. SAM employs advanced segmentation techniques, such as semantic segmentation or instance segmentation, to delineate and generate precise masks outlining the object within the defined ROI. SAM's capability to generate accurate masks around objects of interest serves as a critical pre-processing step within our methodology, providing refined and updated points of interest for subsequent input into tracking algorithms like TAPIR. By harnessing SAM's segmentation capabilities, our methodology aims to improve the robustness and adaptability of SOT in satellite imagery by refining the initial points for enhanced tracking accuracy across frames.

Also, we leverage the TAPIR tracker \cite{tapir} within our methodology. TAPIR operates by initially selecting specific points of interest within the first frame of the video. These points, often representing distinctive features or objects, are tracked across subsequent frames using an optical flow estimation technique. This method analyzes the pixel-level changes between frames, estimating the motion of the selected points. TAPIR integrates per-frame initialization, allowing for precise localization of points in each frame, and employs temporal refinement mechanisms to enhance tracking accuracy. The algorithm dynamically adapts the tracking process based on the local properties of the points being tracked, thus enabling robust and precise tracking even amidst challenging scenarios such as occlusions, scale variations, and background changes. By leveraging this algorithm's adaptability and accuracy in tracking specific points, our methodology aims to achieve resilient and effective tracking within the complexities of satellite imagery.

Our methodology for SOT in satellite videos utilized pre-trained models from the official repositories of SAM and TAPIR, avoiding any additional training on these algorithms.

\subsection{Initialization process}
In the initial step of our methodology (part B of Figure \ref{method-overview}), we employ SAM to generate a segmentation mask for the target object. This process is initiated by prompting SAM with the initial bounding box to derive a precise segmentation, effectively delineating the object within the frame. In case of very small objects, e.g. car, we additionally add a crop and resampling step to aid SAM and consequently TAPIR. Subsequently, to establish a robust set of tracking points, we strategically sample N random points from the generated segmentation mask, ensuring a diverse and representative selection. This approach lays the foundation for initializing an adaptive set of points crucial for the subsequent tracking stages.

\subsection{Point propagation between keyframes}
In the point propagation between keyframes stage, our methodology seamlessly integrates the standard TAPIR pipeline to facilitate the tracking of points from one keyframe to the next. TAPIR excels in this process by leveraging its per-frame Initialization and temporal refinement capabilities. At each keyframe transition, the set of tracked points by TAPIR is efficiently renewed, ensuring continuity and precision throughout the sequence. This entails updating the set of tracked points dynamically, with TAPIR meticulously refining their positions to adapt to changes in the object's appearance, scale, and orientation. The per-frame initialization ensures a precise re-initialization of points in each new keyframe, while the temporal refinement enhances the accuracy of point tracking by accounting for motion dynamics over consecutive frames. The inherent adaptability of TAPIR in handling dynamic scenes and nuanced tracking scenarios guarantees a consistent and precise tracking across successive keyframes, contributing significantly to the accuracy and reliability of the tracking process.

\subsection{Keyframe point generation}
In the keyframe point generation stage (part C of Figure \ref{method-overview}), our methodology strategically harnesses the information propagated by TAPIR to refine and optimize the set of tracking points at each keyframe. For every point within the keyframe, previously propagated through the TAPIR pipeline, we prompt SAM individually. SAM generates segmentation masks for each point, capturing the evolving appearance and features of the tracked object. Subsequently, these individual masks are aggregated into an overlap heatmap through point-wise addition. By selecting the area within the heatmap where the most segmentation masks overlap, we indicate the region of maximum consensus among the points. This region becomes the focal point for a refined set of correctly-tracked points, allowing us to pinpoint the corresponding masks and, consequently, the points from which these masks originated.
 
Building upon this refined set of points, we prompt SAM once again, this time with all the selected points. The objective is to generate a new segmentation mask that precisely delineates the object within the keyframe, leveraging the collective information from the strategically chosen points. To maintain diversity and adaptability, we subsequently employ a random sampling strategy to acquire a fresh set of N points on the newly generated mask. This strategic discarding of previous points ensures continual adaptability and responsiveness to evolving object dynamics, enhancing the overall accuracy and robustness of our tracking methodology. The meticulous interplay of SAM and this keyframe-specific point generation process ensures an optimized and adaptive point set for the ensuing tracking stages, contributing to the methodology's efficacy in handling complex tracking scenarios.

\section{EXPERIMENTAL RESULTS}
\label{results}

VISO dataset \cite{viso}, tailored specifically for evaluating detection and tracking of a single or multiple objects within satellite imagery, encompasses a diverse collection of 47 high-resolution satellite videos featuring numerous small and dense moving objects. These videos depict different environmental settings, encompassing object categories such as cars, airplanes, ships and trains, with varying motion patterns. The dataset comprises 3159 tracklets spanning 1.12 million frames at 10 fps, derived from satellite videos captured by the Jilin-1 satellite constellation at different positions of its orbit. For the dataset's utilization in SOT, tracklets are generated for individual instances in the sequences and ground-truth bounding boxes of objects are provided only in the first frame for initialization, following a standard visual tracking scheme. Compared to existing generic video datasets, the VISO dataset poses challenges with objects having low spatial resolution and a high number of similar instances with variations in point of view, motion blur, and illumination.

We followed the standard evaluation metrics of the VISO dataset, namely Distance Precision Rate (DPR) and Overlap Success Rate (OSR). The first one refers to the percentage of frames whose center location errors are smaller than a given threshold A, while the second one to the percentage of frames whose overlap ratios with the ground-truth box are larger than a given threshold B. We set A=5 pixels due to the small size of objects and B=0.5. Our method was implemented on Pytorch and evaluated on a machine with Ryzen 9 3900X 12 core @3.7GHz CPU, a Nvidia 2080Ti GPU and 64GB RAM. We set the number of keyframes K=20 and the number of sampled points N=20.

\begin{table}
    \centering
    \begin{tabular}{ccc}
        \textbf{Method} & \textbf{DPR (\%)} & \textbf{OSR (\%)} \\ \hline
         SiamRPN++ \cite{siamrpn++} & 47.9 & 19.9 \\
         SiamBAN \cite{siamban} & 48.0 & 18.8 \\
         CFME \cite{cfme} & 50.4 & 28.2 \\
         MCCT \cite{mcct} & 60.5 & 34.7 \\
         ECO \cite{eco} & 61.2 & 34.5 \\
         SRNK \cite{guo2022first} & 75.8 & 33.0 \\
         KYS-ME \cite{guo2022first} & 78.8 & 34.0 \\
         MACF \cite{lin2024motion} & 79.4 & 35.5 \\
         \hline
        \textbf{Ours} & 63.9 & 36.5\\
    \end{tabular}
    \caption{Results of SOT in VISO dataset.}
    \label{results-table}
\end{table}

\begin{figure}[ht!]
\begin{minipage}[b]{0.48\linewidth}
  \centering
    \centerline{\epsfig{figure="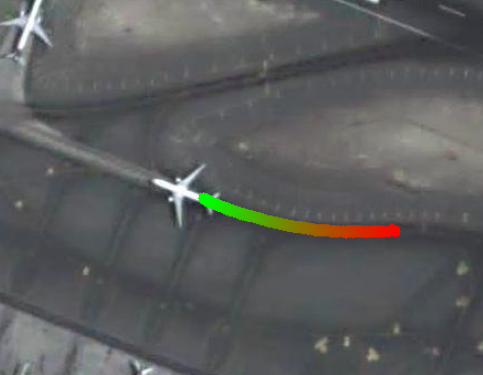",width=3.5cm, height=3cm}}
    \centerline{(a)}\medskip
  
\end{minipage}
\hfill
\begin{minipage}[b]{0.48\linewidth}
  \centering
    \centerline{\epsfig{figure="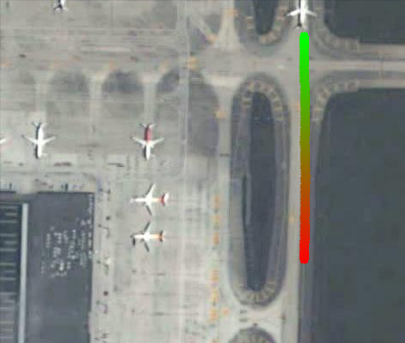",width=3.5cm,height=3cm}}
    \centerline{(b)}\medskip
\end{minipage}
\vfill
\begin{minipage}[b]{0.48\linewidth}
  \centering
    \centerline{\epsfig{figure="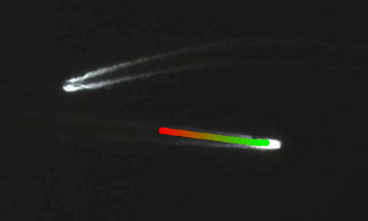",width=3.5cm, height=3cm}}
    \centerline{(c)}\medskip
  
\end{minipage}
\hfill
\begin{minipage}[b]{0.48\linewidth}
  \centering
    \centerline{\epsfig{figure="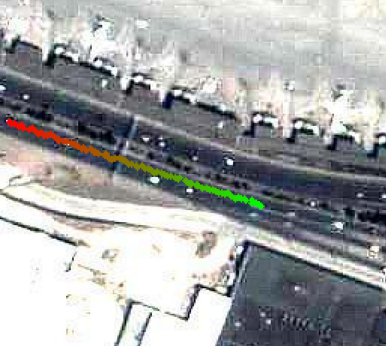",width=3.5cm,height=3cm}}
    \centerline{(d)}\medskip
\end{minipage}

\caption{Qualitative results on the VISO dataset. Red and green color indicate the beggining and the end of the trajectory of a target respectively.}
\label{results-fig}
\vfill
\end{figure}

We compare our method to several trackers that are either specifically designed for satellite data \cite{cfme}, either based on correlation and complex filters \cite{mcct} \cite{guo2022first} \cite{lin2024motion} or Siamese networks \cite{siamrpn++} \cite{siamban}. Our method scores competitive DPR and OSR scores against them, yielding 63.9\% and 36.5\% respectively, as shown in Table \ref{results-table}.
These scores affirm the efficacy of our method in precisely localizing and tracking objects within the challenging context of satellite imagery. For an intuitive and qualitative evaluation, we visualize the tracking results achieved by our method on various sequences of the VISO dataset, as shown in Figure \ref{results-fig}. The visualizations affirm the robust performance of our approach in handling diverse scenes and challenging conditions within satellite imagery.

Our results on this dataset are noteworthy, given the additional merit that our methodology leverages pre-trained models of SAM and TAPIR, without involving any training process. The reliance on these pre-existing, well-performing models contributes to the adaptability of our approach, demonstrating the efficacy of prompt engineering in object tracking without the need for additional training efforts. However, it should be noted that the selection of the hyperparameters K and N is crucial to our method and results may greatly vary for slight disturbances in their optimal values. Overall, our approach is mostly limited by TAPIR's short-term point tracking performance. 

\section{CONCLUSIONS}
\label{sec:majhead}

Our prompt-engineering approach for object tracking in satellite videos, utilizing pre-trained models SAM and TAPIR, has demonstrated remarkable adaptability and showcased strong performance on the VISO dataset, outperforming other satellite trackers. This work emphasizes the potential of prompt-guided solutions, providing an efficient strategy for SOT in satellite imagery. However, our approach encourages further exploration into prompt engineering and future research should be focused on adapting other point trackers and exploring more complex tracking scenarios where swift or unpredictable object movements occur.

\section{Acknowledgements}
\label{sec:acknoledge}

Part of this research was supported by the research project BiCUBES “Analysis-Ready Geospatial Big Data Cubes and Cloud-based Analytics for Monitoring Efficiently our Land \& Water” funded by HFRI (grant: 03943).
\bibliographystyle{IEEEbib}
\bibliography{strings,refs}

\end{document}